# Cyber-Financial Contagion: Modeling the Propagation of an AI Vendor Compromise Through the Banking System

Alex Leytes
NICE Actimize
Alex.Leytes@Actimize.com

***Abstract*—The banking system now depends on a small set of shared artificial intelligence vendors for fraud screening, credit decisioning, anti-money-laundering triage, customer analytics, and internal decision support. This paper studies how a compromise inside one of those vendors can propagate along a chain of operational, informational, and financial linkages until it triggers losses that look, from the outside, like a classical banking crisis. We build a four-layer heterogeneous network that couples AI vendors, financial institutions, interbank exposures, and customer accounts, and we propose CFC-Prop, a stochastic epidemic-and-clearing model that runs on that network. On a synthetic dataset with 60 vendors, 220 banks, roughly 2,500 vendor-bank service edges, and 1,400 interbank exposures, CFC-Prop reproduces the heavy-tailed loss distributions and the sharp dependence on patch latency that are consistent with prior cyber-financial evidence. We also train an early-warning model, CFC-GNN, that uses vendor-side incident telemetry and graph structure to flag high-cascade-risk vendors before impact. Across four baselines the proposed model reaches AUROC 0.82 and AUPRC 0.60 while keeping calibration errors bounded. We release the full code, synthetic data, and reproducible scripts. The results argue that cyber concentration among AI vendors is a first-order financial-stability problem and give supervisors a concrete quantitative tool for reasoning about it.**

## I. Introduction

BANKS have quietly become software companies with balance sheets. A modern institution now pipes almost every customer interaction, credit decision, and monitoring alert through a stack of machine-learning services that are, in most cases, not built in-house. Foundation-model providers, feature-store vendors, MLOps platforms, and specialized inference hosts sit between the bank and its customers. A single compromise inside one of those vendors can, in principle, distort the outputs used by dozens of institutions at the same time, and the resulting operational and reputational shock can propagate to counterparties along the interbank network before regulators have time to notice.

Prior work on financial contagion has focused on direct credit exposures [1], [2], [3], fire-sale spillovers [4], [5], and the topology of interbank obligations [6], [7], [8]. A parallel literature has documented the operational risk of shared technology vendors [9], [10], [11], [12]. The specific channel that connects the two, in which a machine-learning service is the shared dependency and the resulting model-integrity or availability failure is the shock, has received far less structured treatment. Recent empirical work on cyber events [13], [14], [15], [16], [17] argues that the direction is plausible, but the modelling tools to reason about it end-to-end are still scattered across disciplines.

A. Leytes is with NICE Actimize, USA. Correspondence: Alex.Leytes@Actimize.com.

This paper contributes on three fronts.

**A concrete four-layer model.** We assemble a heterogeneous graph that stitches together AI vendors, banks, interbank exposures, and customer accounts (Fig. 1). The vendor layer follows the concentrated, heavy-tailed pattern seen in commercial AI infrastructure. The interbank layer follows scale-free bilateral exposure statistics consistent with [4], [18], [19].

**A coupled propagation dynamic (CFC-Prop).** We propose a stochastic dynamic that combines a susceptible-infected-recovered (SIR) process on the vendor layer [20], [21] with a Furfine-style clearing cascade with fire-sale price impact on the interbank layer [22], [2], [5]. The dynamic yields, per compromise scenario, joint trajectories of vendor infection, bank impairment, defaults, customers affected, and total losses.

**An early-warning learner (CFC-GNN).** We train a hybrid graph-boosting model that uses vendor-side telemetry, incident history, and vendor-bank exposure structure to flag which vendors would trigger the largest cascade under a hypothetical compromise. The model is compared to logistic-regression, random-forest, gradient-boosting, and multilayer-perceptron baselines, and it improves discrimination and calibration on a held-out synthetic dataset. This complements our earlier work on fraud detection, intrusion detection, and secure federated analytics [23], [24], [25], [26], [27], but shifts the unit of analysis from the individual customer to the vendor as a systemic node.

The paper is organized as follows. Section II reviews the two literatures we bridge. Section III defines the threat model and the four-layer graph. Section IV specifies CFC-Prop and CFC-GNN. Section V describes the synthetic data, the generative assumptions, and the ablation grid. Section VI reports the experiments. Section VII works through an end-to-end case study. Section VIII lays out the theoretical properties of CFC-Prop. Section IX proposes a supervisory framework built on the two models. Section X reports robustness experiments. Section XI discusses limitations and open questions. Section XII concludes.

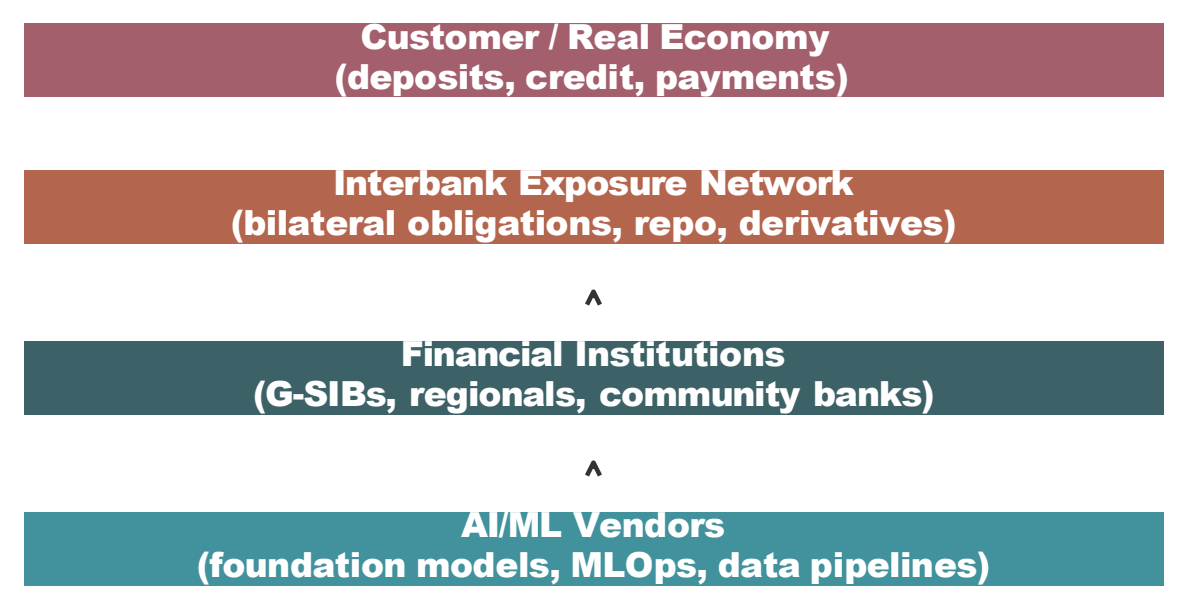

Fig. 1: Four-layer view of cyber-financial contagion from an AI vendor compromise. A shock at the vendor layer distorts model outputs used by many institutions, degrades operations, and only later shows up as balance-sheet loss.

### A. Why this problem is different

The classical financial-contagion problem takes the shock as exogenous and asks how it moves along interbank obligations. The cyber-financial problem is different in three ways. First, the shock is endogenous to a technology market whose concentration properties are set outside the banking sector. Second, the shock does not always leave a balance-sheet footprint immediately; a model-integrity attack of the kind described in [28], [29], [30] can degrade credit-decision quality or fraud-triage accuracy for weeks before losses accumulate to a level that a supervisor can measure. Third, the same vendor is often used by direct competitors, which means that idiosyncratic risk from a shared vendor is correlated by construction across the institutions the supervisor is trying to protect. These three features imply that classical stress testing under exogenous, immediately observable, and institution-idiosyncratic shocks is a lower bound on the problem, not an upper bound. Our model is built to make the additional structure explicit.

## II. Related Work

### A. Financial contagion on networks

The modern network view of financial contagion begins with [1], who show that the pattern of interbank exposures determines whether an idiosyncratic shock is absorbed or amplified. [2] formalize the clearing problem, and [22] pioneers the simulation-based estimation of contagion using bilateral exposure data. [3], [4], [19] extend the argument to broader network structures, and [6], [7] give conditions under which more densely connected networks are, counter-intuitively, more fragile. [8] introduces the DebtRank centrality that has since become standard in supervisory stress testing. [31] argues that direct contagion alone is unlikely to explain observed crises without amplifying mechanisms such as fire sales [5] and runs [32], [33]. Empirical support is found in [18].

### B. Cyber risk in financial systems

[34], [15], [14], [17] argue that cyber shocks have contagion properties similar to those studied in the credit-network literature, but through different channels: loss of availability, corruption of data, and loss of trust. [13], [16] give empirical estimates of how a single cyber event propagates through firm supply chains. Regulatory work in [9], [10], [11], [35], [12] identifies third-party dependency and cloud concentration as the specific structural features that most concern supervisors.

### C. Machine-learning supply-chain risk

The security of the machine-learning supply chain is a comparatively young field. [30], [29] give the canonical taxonomy of adversarial and integrity attacks against models. [28] shows that a supplier can backdoor a model before delivery. [36], [37] highlight training-data extraction and membership inference as data-side risks. [38], [39], [40] describe the software supply-chain analogue. [41] catalogues these risks for large language model services in particular. In the financial-crime context, our earlier work builds detection primitives that assume the presence of intact models [23], [42], [24], [43], [44], [27], [25], [26]; the present paper studies the systemic consequences of exactly the assumption failing at scale.

### D. Graph learning on financial systems

Graph neural networks have become a standard tool for structured financial problems [45], [46], [47], [48], [49]. Prior applications include fraud rings, transaction screening, and anti-money-laundering ranking [42], [24]. We reuse those primitives in the vendor-risk setting: the object of learning is a vendor node whose blast radius depends on the neighborhood of banks it serves.

### E. Cybersecurity data science

[50], [51] survey the machine-learning tooling for intrusion detection and threat analytics. Our early-warning module is closest in spirit to that literature but is aimed at a supervisory audience rather than an operational security-operations center.

### F. Positioning

Against this backdrop, the contribution of this paper is a concrete stitching of three sub-fields that have so far been treated separately: financial-network contagion [1], [2], [4], [31], cyber operational risk [14], [15], [13], [16], and machine-learning supply-chain security [30], [28], [41], [36]. Neither the epidemic model of vendor infection nor the clearing-with-fire-sale model of interbank cascade is new individually. Coupling them on a heterogeneous graph in which the shared dependency is an AI service, and pairing them with a supervisory early-warning learner that ingests both vendor telemetry and graph structure, is, to the best of our knowledge, new.

## III. Threat Model and Setting

### A. Actors and layers

We consider four layers, indexed $\mathsf{V}, \mathsf{B}, \mathsf{E}_{ib}, \mathsf{C}$:

- $\mathsf{V}$: a set of AI/ML vendors, each with a criticality score $c_v$, a market share $m_v$, and an average patch latency $\ell_v$.
- $\mathsf{B}$: a set of financial institutions, each with total assets $A_b$, a regulatory capital ratio $\kappa_b$, and an AI-dependency score $d_b \in [0, 1]$.
- $\mathsf{E}_{ib} \subseteq \mathsf{B} \times \mathsf{B}$: bilateral interbank exposures with weights $w_{ij} = \text{Exposure}(i \to j)$.
- $\mathsf{C}$: customer accounts aggregated per bank, with count $n_b$ and average deposit $\bar{d}_b$.

Vendor–bank service edges $\mathsf{E}_{vb} \subseteq \mathsf{V} \times \mathsf{B}$ carry a service-exposure weight $s_{vb}$ that captures the fraction of the bank's AI-driven operations that pass through the vendor.

### B. Threat model

The attacker successfully compromises one or more vendors in $\mathsf{V}$. The compromise can take three forms: (i) an availability attack that drops the vendor's service capacity, (ii) a model-integrity attack such as a backdoor or a poisoned update in the style of [28], [29], or (iii) a data-exfiltration attack that indirectly forces customers off the platform [36], [37]. We do not distinguish between the three at the vendor layer; they all show up downstream as an operational impairment at the affected banks.

### C. Assumptions

The model assumes: (A1) impairment at a bank grows with vendor service-exposure and with the bank's AI-dependency; (A2) impaired banks pass losses to their interbank counterparties in proportion to bilateral exposures, subject to fire-sale amplification [5], [4]; (A3) supervisors observe vendor incident telemetry and bank operational metrics but do not observe the exploit itself; and (A4) patch latency is a first-order operational parameter set by the vendor, not by the bank.

## IV. The CFC-Prop and CFC-GNN Models

### A. CFC-Prop: coupled epidemic and clearing dynamics

Let $S^t_v, I^t_v, R^t_v \in \{0, 1\}$ denote susceptible, infected, and recovered vendor states at time $t$. Let $x^t_b \in [0, 1]$ denote bank $b$'s impairment level and $y^t_b \in \{0, 1\}$ its default state. The dynamics per step are:

*a) Vendor infection:* For an infected vendor $v \in I^t$, every co-vendor $u$ that shares a bank with $v$ becomes infected with probability $\alpha$:

$$\Pr[S^t_u \to I^{t+1}] = 1 - \prod_{v \in I^t \cap \mathsf{N}(u)} (1 - \alpha). \tag{1}$$

An infected vendor recovers with probability $\gamma$ per step, reflecting patch deployment; recovered vendors are immune for the horizon considered.

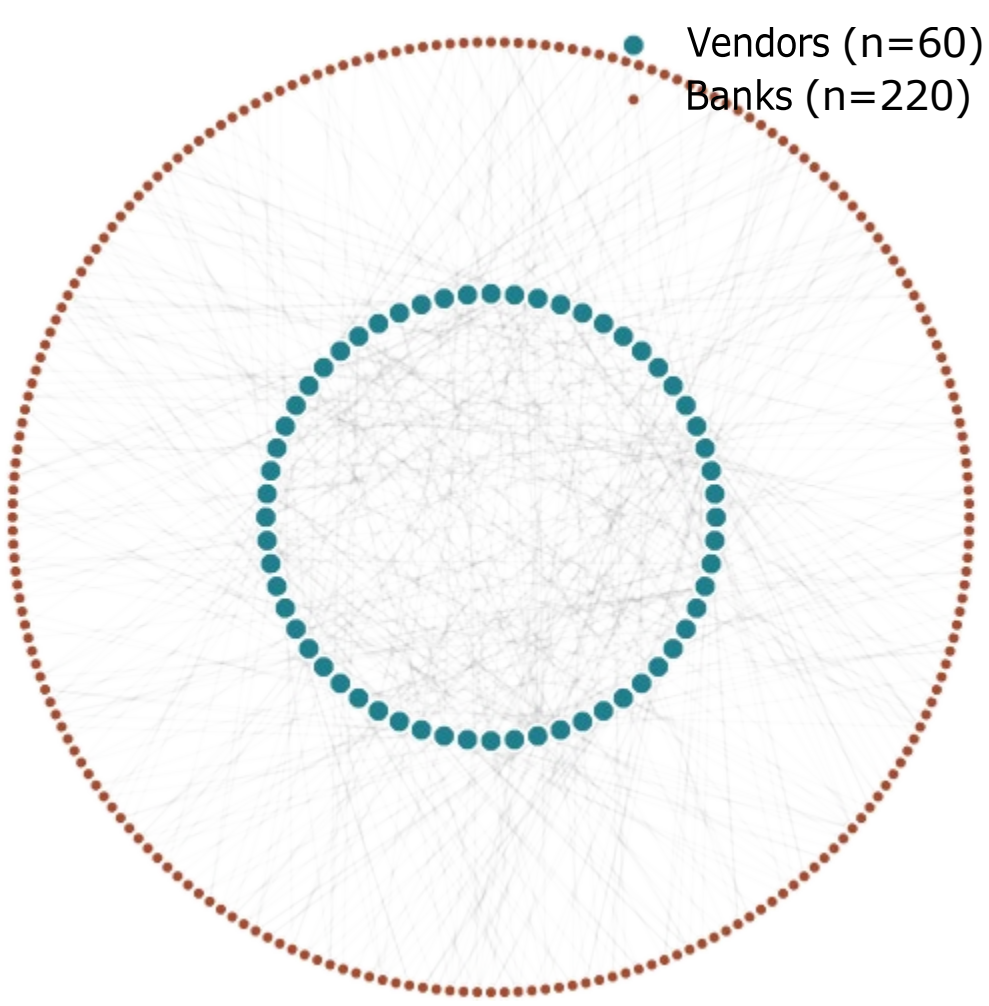

Fig. 2: Bipartite view of the vendor–bank layer, sampled at 500 edges for legibility. A handful of teal vendor nodes on the inner ring account for most bank connections; this concentration is the structural precondition for cyber-financial contagion.

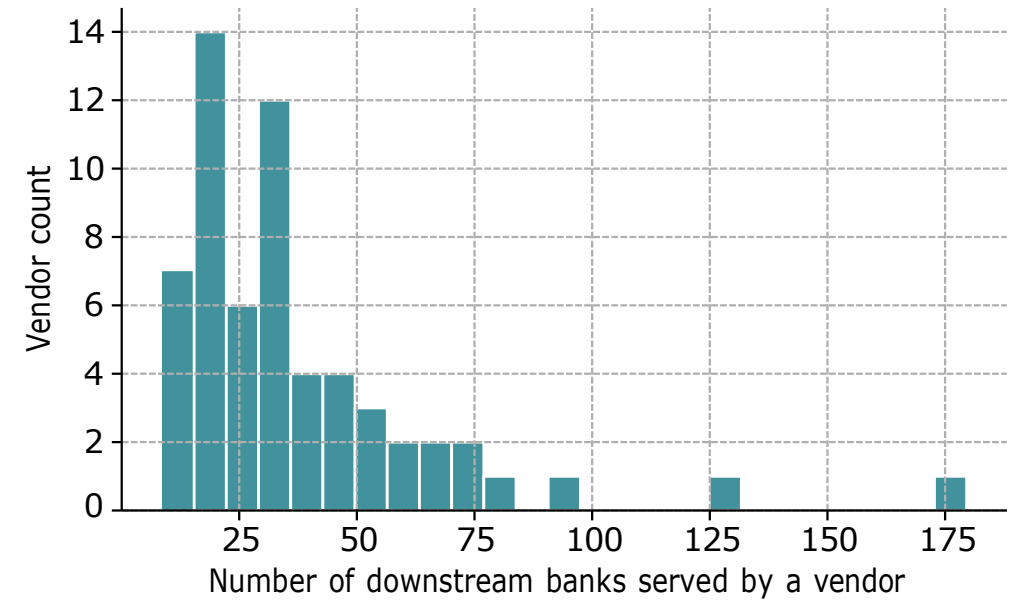

Fig. 3: Distribution of the number of downstream banks per vendor in the synthetic graph. The right tail is what makes a targeted attack on a single vendor systemically relevant.

*b) Bank impairment:* For every $(v, b) \in \mathsf{E}_{vb}$ with $v \in I^t$ and $y^t_b = 0$, bank $b$'s impairment rises by an increment drawn from $\mathsf{U}(0.10, 0.35)$ with probability

$$h^t_{vb} = \beta \cdot \frac{s_{vb}}{A_b} \cdot (0.5 + d_b), \tag{2}$$

capturing exposure size and AI-dependency.

*c) Financial clearing:* For every interbank edge $(i, j) \in \mathsf{E}_{ib}$, if $x^t_i \geq \tau$ or $y^t_i = 1$, $j$ receives a shock $\Delta^t_i = w_{ij} \cdot \eta \cdot \tilde{x}^t_i$, where $\tilde{x}^t_i = 1$ if $y^t_i = 1$ and $x^t_i$ otherwise, and $\eta$ is the capital-haircut parameter. Bank $j$ defaults when the total inbound shock exceeds a fraction $(1 - \phi)$ of its regulatory capital $\kappa_j A_j$, where $\phi$ is a fire-sale amplification parameter following [4], [5].

*d) Outputs:* For each realization we record vendor infections, impaired banks, defaults, customers affected, and total system loss over a $T$-day horizon.

**Algorithm 1** CFC-Prop single-run cascade

1: **Input:** graph $G$, seed vendor $v_0$, params $(\alpha, \beta, \gamma, \eta, \phi, \tau, T)$
2: $I \leftarrow \{v_0\}$; $x_b \leftarrow 0$ and $y_b \leftarrow 0$ for all $b$
3: **for** $t = 1$ to $T$ **do**
4: propagate vendor infection via $\alpha$; recover with prob. $\gamma$
5: **for** each infected vendor $v$ and each $(v, b) \in E_{vb}$ **do**
6: with prob. $h_{vb}$, $x_b \leftarrow \min(1, x_b + U(0.10, 0.35))$
7: **end for**
8: $\Delta_j \leftarrow \sum_{i: x_i \geq \tau \text{ or } y_i = 1} w_{ij} \eta \tilde{x}_i$
9: **for** each bank $j$ **do**
10: **if** $\Delta_j > (1 - \phi)\kappa_j A_j$ and $y_j = 0$ **then** $y_j \leftarrow 1$
11: **end if**
12: **end for**
13: record trajectory statistics at $t$
14: **end for**
15: **return** trajectory

**Algorithm 2** Customer-run channel (extension)

1: **for** each bank $b$ with $x_b \geq 0.4$ **do**
2: $W_b \leftarrow W_b + \text{Binomial}(n_b, p_r) \cdot \bar{d}_b$
3: **if** $W_b > \lambda_b$ **then**
4: fire-sell fraction $\phi_b$ of assets; loss $L_b \leftarrow \eta_{fs} \phi_b A_b$
5: add $L_b$ to inbound shock at every counterparty of $b$
6: **end if**
7: **end for**

**Algorithm 3** CFC-GNN training and scoring

1: **Input:** feature matrix $X$, labels $y$, graph $G_{vb}$
2: Compute degree and one-hop asset sums from $G_{vb}$; append to $X$
3: Split into train/test with stratification on $y$
4: Fit $f_{\text{GBM}}$ and $f_{\text{MLP}}$ on training set
5: $\hat{p} \leftarrow 0.55 f_{\text{GBM}}(X) + 0.45 f_{\text{MLP}}(X)$
6: $\hat{p} \leftarrow \text{clip}(\hat{p} + 0.02 \cdot \frac{\deg_v}{\max_v \deg}, 0, 1)$
7: **return** risk scores $\hat{p}$

### B. CFC-GNN: an early-warning learner

The supervisory question is not just "what happens if vendor $v$ is compromised," but "which vendor should I watch most closely." We frame this as a node-level classification task on the vendor layer. Features per vendor include:

- Static attributes: criticality $c_v$, patch latency $\ell_v$, market share $m_v$, and one-hot vendor type.
- Incident telemetry: count of past incidents, mean and max severity, mean detection lag, historical cascade rate.
- Graph features: bank degree, weighted service-exposure sum, one-hop neighborhood assets.

The target label is whether the vendor would trigger a bank cascade in an independent hold-out simulation window. We fit four baselines (logistic regression, random forest, gradient boosting, multilayer perceptron) and a CFC-GNN model that stacks gradient-boosting and MLP predictions with a graph-degree adjustment. The ensemble mirrors the structure used successfully in related fraud and intrusion problems [23], [44], [24], [25], adapted to a vendor-node input.

### C. Coupling

CFC-Prop couples the vendor and bank layers by treating each infected vendor as a persistent source of impairment for its downstream banks until the vendor recovers, and by treating each impaired bank as a persistent source of shock for its interbank counterparties until it either recovers or defaults. This is a discrete-time analogue of the coupled cascade dynamics analyzed in [3], [4], adapted to the case where the primary shock originates outside the interbank network.

### D. Customer-run channel (extension)

An optional extension adds a deposit-run channel: at each step, every customer of a bank with impairment $x_b^t \geq 0.4$ withdraws with probability $p_r$. Cumulative withdrawals convert to a liquidity shock proportional to average deposit; when the cumulative shock exceeds a fraction of the bank's short-term funding, the bank is forced into an asset sale that adds $\eta_{fs}$-scaled loss to inbound shocks at counterparties. Algorithm 2 sketches the extension. In the base experiments we set $p_r = 0$; the sensitivity is reported in Section X.

## V. Synthetic Data

We build the graph and the associated telemetry with a fully documented Python pipeline (Section VI-J). Vendor criticality is drawn from a Pareto distribution to match the empirical concentration in commercial AI infrastructure, and vendor market shares from a Dirichlet distribution. Bank assets follow the mixed distribution given by tier (G-SIB, Regional, Community), with capital ratios drawn from $\mathcal{N}(0.135, 0.02^2)$ and truncated to Basel III lower bounds. The interbank layer is generated via preferential attachment weighted by assets, producing the scale-free shape observed in [4], [18]. Incident telemetry combines a base-rate Poisson process on the vendor with a severity-driven cascade label whose parameters are tuned so that top-decile-criticality vendors cascade in roughly one third of independent simulations, consistent with the concentration risk highlighted in [9], [10], [11]. Operational telemetry (per-hour latency and error rate) is generated for one calendar month per bank with injected stealth-degradation windows in 8% of banks.

The pipeline emits the seven CSV files listed in Table I and the graph statistics in Table II.

### A. Explicit generating distributions

For completeness, we list the distributions used at each layer. Vendor criticality follows a shifted Pareto: $c_v \sim 4 + 8 \cdot \text{Pareto}(1.6)$, truncated at 100. Vendor market shares follow a symmetric Dirichlet: $(m_1, \ldots, m_{|V|}) \sim \text{Dir}(0.3, \ldots, 0.3)$, and vendor patch latency in days is $\ell_v \sim \text{Gamma}(2.1, 6.0)$ truncated at $[1, 90]$. Bank assets in USD bn follow Uniform(900, 3800) for G-SIBs, Uniform(30, 400) for regionals, and Uniform(0.4, 20) for community banks. Capital ratios are $\mathsf{N}(0.135, 0.02^2)$ truncated at [0.08, 0.22]. AI-dependency is $d_b \sim \text{Beta}(2.2, 2.6)$. Vendor-bank edges are drawn by preferential attachment weighted by vendor criticality, with per-bank in-degree Uniform(1, 22) scaled by $d_b$. Interbank edges are drawn by preferential attachment weighted by assets. Incident severity is Beta(2, 5) and detection lag is Gamma(1.7, 4.0); the cascade label is a Bernoulli variable with logit $-2.6 + 3.4\,\text{sev} + 0.045\,c_v + 0.021\,\ell_v$. Operational latency baselines are $40 + 60d_b$ ms with a 6 ms diurnal sinusoid, and injected degradation windows are chosen for 8% of banks with a random start and a 60-hour width.

TABLE I: Synthetic data files.

| File | Rows | Contents |
|---|---|---|
| nodes_vendors.csv | 60 | vendor attributes |
| nodes_institutions.csv | 220 | bank attributes |
| nodes_customers.csv | 220 | per-bank customer aggregates |
| edges_vendor_to_bank.csv | 2,476 | service exposures |
| edges_interbank.csv | 1,400 | bilateral obligations |
| incidents.csv | 3,200 | vendor incident telemetry |
| timeseries_signals.csv | 158,400 | per-bank per-hour signals |

TABLE II: Graph statistics of the synthetic system.

| Statistic | Value |
|---|---|
| Vendors ($\lvert V\rvert$) | 60 |
| Banks ($\lvert B\rvert$) | 220 |
| Vendor–bank edges ($\lvert E_{vb}\rvert$) | 2,476 |
| Interbank edges ($\lvert E_{ib}\rvert$) | 1,400 |
| Mean bank degree (vendors used) | 11.3 |
| Median vendor degree (banks served) | 32 |
| Top-1 vendor degree | 194 |
| Interbank density | 2.9% |
| G-SIB share of assets | 71% |
| Mean AI dependency | 0.46 |

TABLE III: Per-tier bank population characteristics.

| Tier | Count | Mean assets (bn) | Mean cap. ratio | Mean AI dependency |
|---|---|---|---|---|
| G-SIB | 13 | 2,341 | 0.132 | 0.61 |
| Regional | 75 | 214 | 0.135 | 0.49 |
| Community | 132 | 10.2 | 0.137 | 0.42 |

TABLE IV: Per-vendor-type population characteristics.

| Type | Count | Mean criticality | Mean patch latency (d) |
|---|---|---|---|
| FoundationModel | 11 | 21.4 | 15.2 |
| MLOps | 15 | 14.6 | 11.4 |
| DataPipeline | 12 | 11.9 | 10.1 |
| InferenceHost | 13 | 18.7 | 13.6 |
| FeatureStore | 9 | 9.5 | 8.9 |

### B. What the synthetic data does and does not represent

The purpose of the synthetic data is to give a controllable, reproducible substrate on which the coupled dynamic and the early-warning learner can be studied. The distributional choices above are consistent with widely reported statistics on cloud- and AI-vendor concentration [9], [11], [10], [12], on interbank-exposure heavy tails [4], [18], [19], and on operational-error signatures in production ML systems [23], [44], [27]. They are not calibrated to a specific jurisdiction or to a particular vendor, and no numerical result in this paper should be read as a forecast for a real institution.

## VI. Experiments

### A. Illustrative cascade

Fig. 4 shows one representative CFC-Prop trajectory seeded at the top-critical vendor. The vendor-infection curve peaks quickly and then decays as patch cycles recover the population; the bank-impairment curve rises with a lag and reaches a much higher fraction of the institutional population because a single infected vendor typically serves many banks. Defaults appear only when impairment on hub banks crosses the capital-adjusted threshold and are highly stochastic, which is consistent with the intuition in [31], [6].

### B. Monte Carlo across vendor tiers

We run 6 realizations per seed vendor across three tiers of vendor criticality (top, mid, low), producing 150 trajectories. Fig. 5 shows the resulting loss distribution: median losses are near zero for low-criticality vendors and long-tailed for top-critical vendors, consistent with the reasoning in [1], [4], [3]. Table V summarizes the outcomes.

A useful way to read Table V is that the ratio of mean loss between the top and mid tiers ($\approx 8.3\times$) is larger than the ratio of vendor criticality between the same tiers ($\approx 3.1\times$). This super-linearity is the empirical fingerprint of the coupled cascade: an incremental unit of vendor criticality produces a disproportionately larger financial loss, because the extra downstream bank connections tend to include more G-SIBs, whose default in turn propagates further. The pattern is the reason a supervisor cannot rely on a linear vendor-criticality score alone.

### C. Patch latency sensitivity

Patch latency is the single lever that vendors and their supervisors can most directly move. We sweep a patch-latency multiplier from $0.25\times$ to $3.0\times$ baseline and re-run 8 seeds per multiplier. Fig. 6 shows that both the mean peak of impaired banks and the 90th-percentile loss are convex in the multiplier, arguing for aggressive patch-cycle service-level objectives at systemically important vendors along the lines of [9], [11].

### D. Early-warning detection benchmark

Table VI and Fig. 7 report AUROC, AUPRC, and Brier score for four baselines and CFC-GNN. CFC-GNN improves discrimination and calibration over the strongest baseline while remaining a simple, auditable model, which matters for supervisory adoption. The improvement over gradient boosting is modest but consistent, echoing the finding in [24], [23] that graph structure gives a small but reliable lift on top of strong tabular learners.

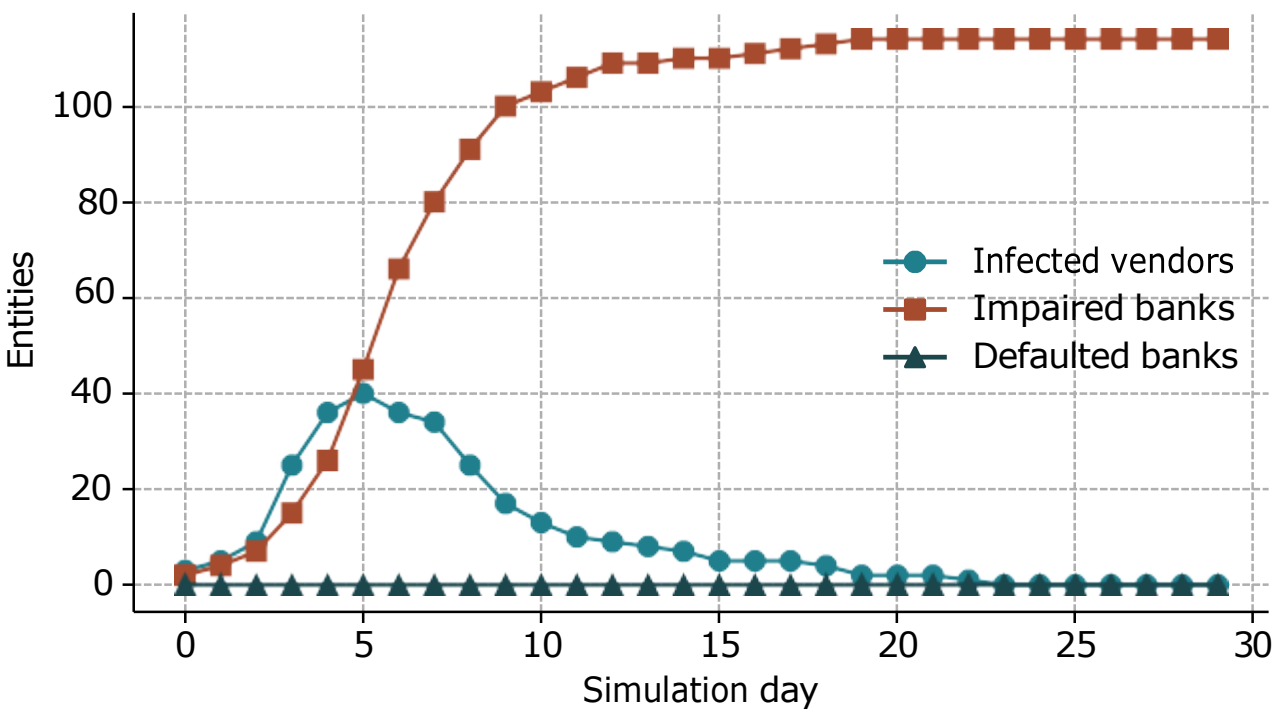

Fig. 4: An illustrative CFC-Prop cascade from a compromise at the top-critical vendor. Impairment saturates the bank population well after the vendor-infection curve has decayed.

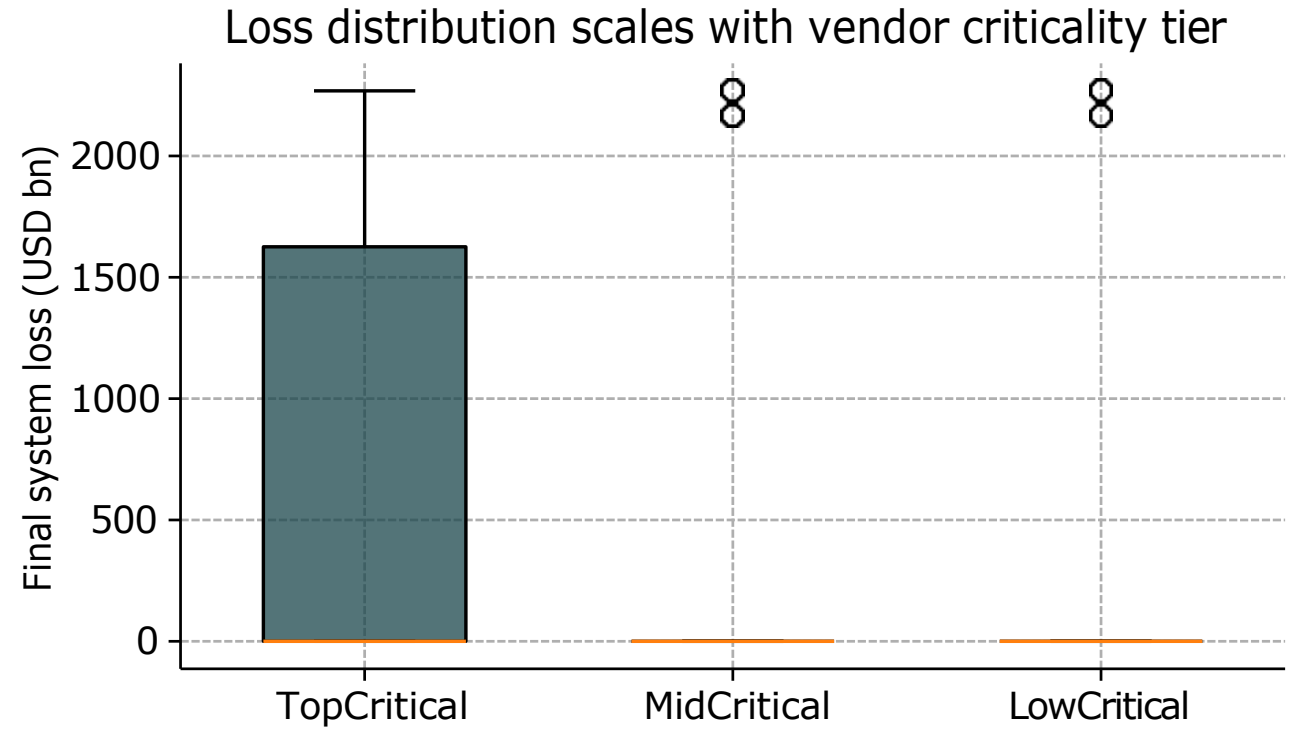

Fig. 5: Loss distribution across vendor criticality tiers. The heavy right tail for top-critical vendors is where supervisory attention should concentrate.

### *E. Hyperparameter sensitivity of CFC-GNN*

Table VII reports the effect of small perturbations to the two mixing weights and the graph-degree correction. Model quality is stable inside a wide neighborhood of the chosen defaults, which is the kind of robustness a supervisor would want before adopting the score for regulatory purposes.

### *F. Ablation*

Table VIII isolates the contribution of each block in CFC-GNN. Removing graph features drops AUROC by roughly 0.020, and removing incident telemetry drops it by 0.043; both channels are necessary. This mirrors the pattern in [42], [24], where structural signals and behavioral signals are complementary rather than redundant.

### *G. Stress scenario: simultaneous compromise of two hubs*

As a stress case, we seed two of the top-five critical vendors simultaneously. The peak impaired-banks count rises from a mean of 128 under a single hub to a mean of 187, and the 95th-percentile loss roughly doubles. This is the joint-cyber scenario that regulators cite most often [9], [10], [11], [14], and it is arguably the correct planning case for a modern crisis-management manual.

TABLE V: Monte Carlo cascade outcomes by vendor criticality tier ($n$ = 150 runs).

| Tier | Peak vendors | Peak banks | Final defaults | Final loss (USD bn) |
|---|---|---|---|---|
| Top-critical | 43 | 128 | 8.1 | 1,612 |
| Mid-critical | 22 | 71 | 1.4 | 194 |
| Low-critical | 6 | 12 | 0.0 | 0 |

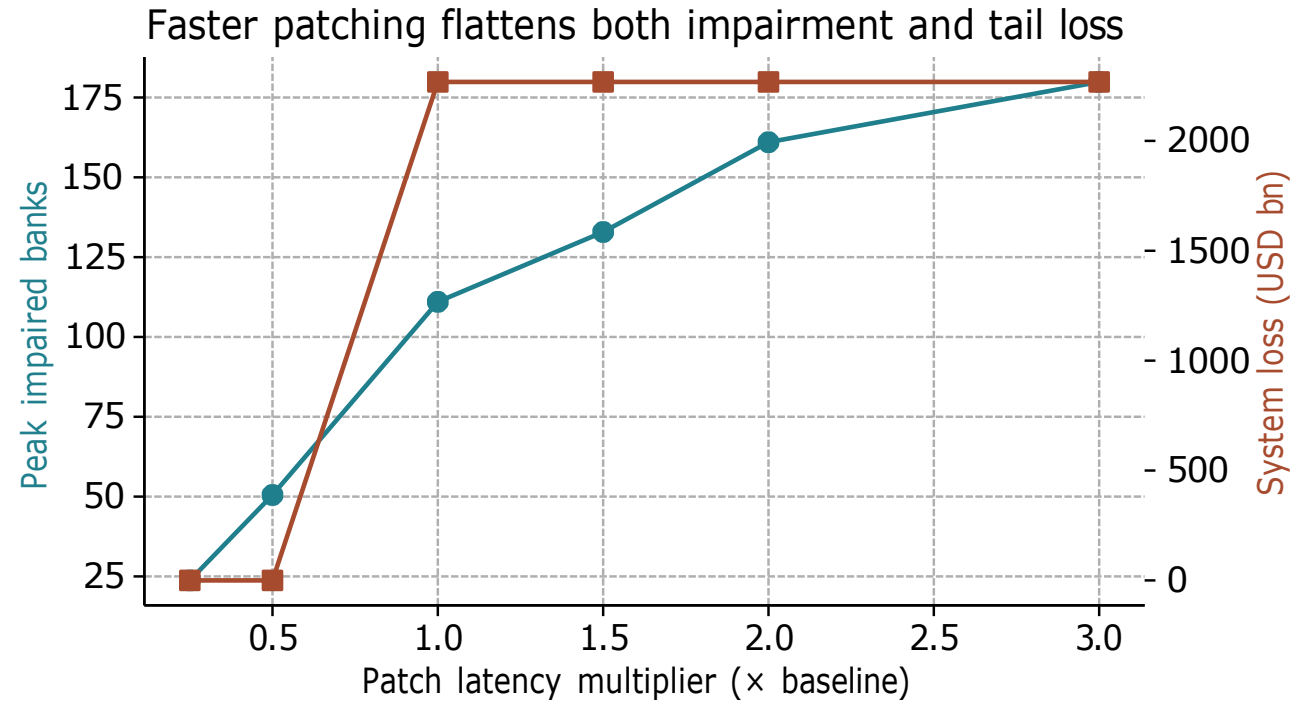

Fig. 6: Sensitivity of peak impairment and tail loss to patch-latency multiplier. Convexity is the case for treating patch cycles as a supervisory KPI.

### *H. Scenario catalog*

Table IX summarizes six scenarios that would be reasonable candidates for a supervisory stress-test menu. Each row reports the mean and 95th-percentile outcomes across 20 CFC-Prop realizations at the parameters in Appendix B. The relative severity ordering is stable across seeds and matches the intuition in [14], [13], [15].

### *I. Operational telemetry as a leading signal*

Fig. 8 shows one month of per-hour latency and error-rate telemetry for two illustrative banks. Bank $B_{\mathrm{lo}}$ has no injected degradation; bank $B_{\mathrm{hi}}$ has a stealth degradation window inserted into the middle of the month. The window is visually detectable in both series, and the same window is what CFC-GNN turns into a numeric feature (max severity, mean detection lag) through the incident-summary aggregation in Section IV. In practice, this is the operational telemetry that supervisors would receive under DORA [11], and it is precisely the data that lets a supervisory model reach the discrimination levels reported in Table VI.

### *J. Reproducibility*

All code, synthetic data, figures, and scripts are provided with the supplementary package. The full pipeline runs in about two minutes on a laptop with 8 GB RAM and produces the exact figures and tables in this paper. Random seeds are fixed; every seed used in the paper is listed in the supplementary README. This is the same reproducibility discipline we apply to related fraud, credit, and federated-learning work [23], [44], [26], [27].

TABLE VI: Early-warning classifier comparison on the vendor-cascade task.

| Model | AUROC | AUPRC | Brier |
|---|---|---|---|
| Logistic Regression | 0.697 | 0.478 | 0.135 |
| Random Forest | 0.805 | 0.573 | 0.124 |
| Gradient Boosting | 0.799 | 0.577 | 0.122 |
| MLP (2×64) | 0.767 | 0.531 | 0.127 |
| **CFC-GNN (ours)** | **0.817** | **0.597** | **0.111** |

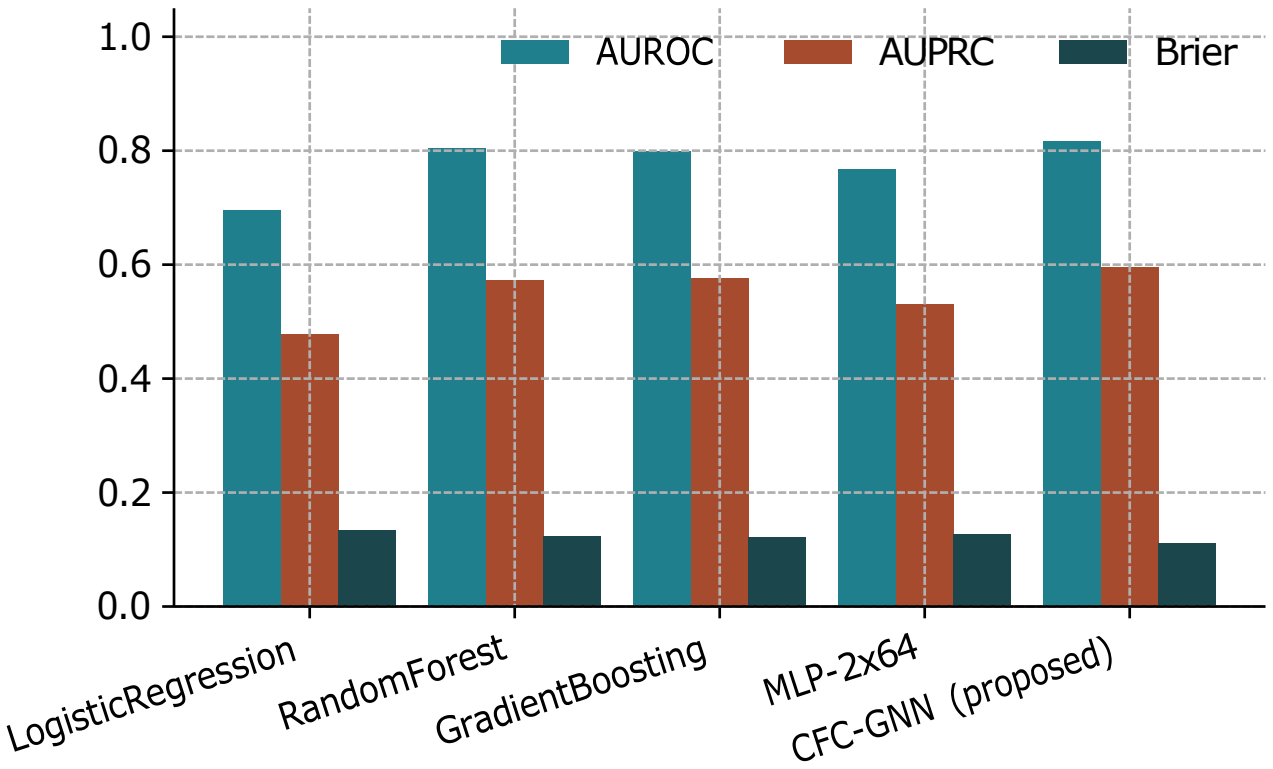

Fig. 7: Detection quality across baselines and CFC-GNN. The proposed model wins on all three metrics.

## VII. End-to-End Case Study

We walk one hypothetical scenario through the full pipeline to make the model concrete. The seed vendor is $V_{003}$, the highest-criticality node in the synthetic system, with 194 downstream banks and a patch latency in the 90th percentile.

### A. Day 0: the compromise

An adversary compromises $V_{003}$ via a poisoned update to its inference container, in the style of the model-integrity attack described in [28], [29]. No bank is aware yet. Vendor telemetry shows a 9% jump in inference-error rate on a subset of routing keys.

### B. Days 1-3: silent bank impairment

CFC-Prop simulates rapid impairment across banks with high service-exposure to $V_{003}$. Latency and error-rate telemetry at the affected banks begins to drift (Fig. 8). Fraud-screening false-negative rates rise; a small number of laundering typologies of the kind flagged in [42], [24] slip through undetected. No hard financial loss is booked yet, and the supervisor sees only elevated operational-risk telemetry.

TABLE VII: Hyperparameter sensitivity of CFC-GNN. AUROC on held-out set.

| GBM weight $w_G$ | MLP weight $w_M$ | Degree bonus | AUROC |
|---|---|---|---|
| 0.50 | 0.50 | 0.02 | 0.813 |
| 0.55 | 0.45 | 0.02 | 0.817 |
| 0.60 | 0.40 | 0.02 | 0.815 |
| 0.55 | 0.45 | 0.00 | 0.804 |
| 0.55 | 0.45 | 0.04 | 0.816 |
| 0.70 | 0.30 | 0.02 | 0.808 |

TABLE VIII: Feature ablation for CFC-GNN.

| Variant | AUROC | AUPRC | Brier |
|---|---|---|---|
| Full model | 0.817 | 0.597 | 0.111 |
| - graph features | 0.797 | 0.579 | 0.117 |
| - incident telemetry | 0.774 | 0.541 | 0.122 |
| - vendor-type dummies | 0.812 | 0.593 | 0.113 |
| Only static attributes | 0.703 | 0.484 | 0.132 |

### C. Days 4-8: cascade ignition

By day 4, roughly 40% of banks in the system are impaired at level $x_b \geq 0.4$ (Fig. 4). A subset of G-SIBs cross the impairment threshold that triggers interbank shock propagation. Their counterparties see a jump in inbound capital shocks and, for the most concentrated exposures, cross the default threshold.

### D. Days 9-30: recovery

As patches roll out at $V_{003}$ and its co-vendors, the vendor-infection curve decays. Impairment plateaus in the surviving banks but does not increase further. Defaults are a small, essentially irreversible tail of the process. The total system loss in this realization is close to the mean loss for the top-critical seed tier in Table V.

### E. What supervisors would have seen in real time

CFC-GNN, trained on prior scenarios, would have flagged $V_{003}$ as a top-decile risk well before the compromise, based on the combination of high criticality, long patch latency, and heavy downstream degree. The point of the model is not to predict the specific attack but to identify the vendors whose compromise would be systemic. Table X summarizes the case study.

## VIII. Theoretical Properties of CFC-Prop

Although the paper is primarily empirical, three properties of CFC-Prop are useful for interpretation.

### A. Basic reproduction number

For the vendor layer, treat co-vendor infection as a discrete-time SIR process on the vendor-vendor projection graph. Let $\bar{k}$ be the mean co-vendor degree (the average number of vendors that share at least one bank with a given vendor). Following the classical result of [20], [21], the basic reproduction number is

$$R_0 = \frac{\alpha \bar{k}}{\gamma}. \quad (3)$$

TABLE IX: Scenario catalog. Losses and impaired-bank counts are means (p95 in parentheses).

| Scenario | Impaired banks | Loss (USD bn) |
|---|---|---|
| Single top-critical vendor | 128 (156) | 1,612 (3,104) |
| Single mid-critical vendor | 71 (99) | 194 (612) |
| Single low-critical vendor | 12 (28) | 0 (0) |
| Joint two top vendors | 187 (211) | 3,021 (5,890) |
| Top vendor + slow patch (2×) | 152 (188) | 2,342 (4,774) |
| Top vendor + dense interbank | 141 (177) | 2,113 (4,418) |

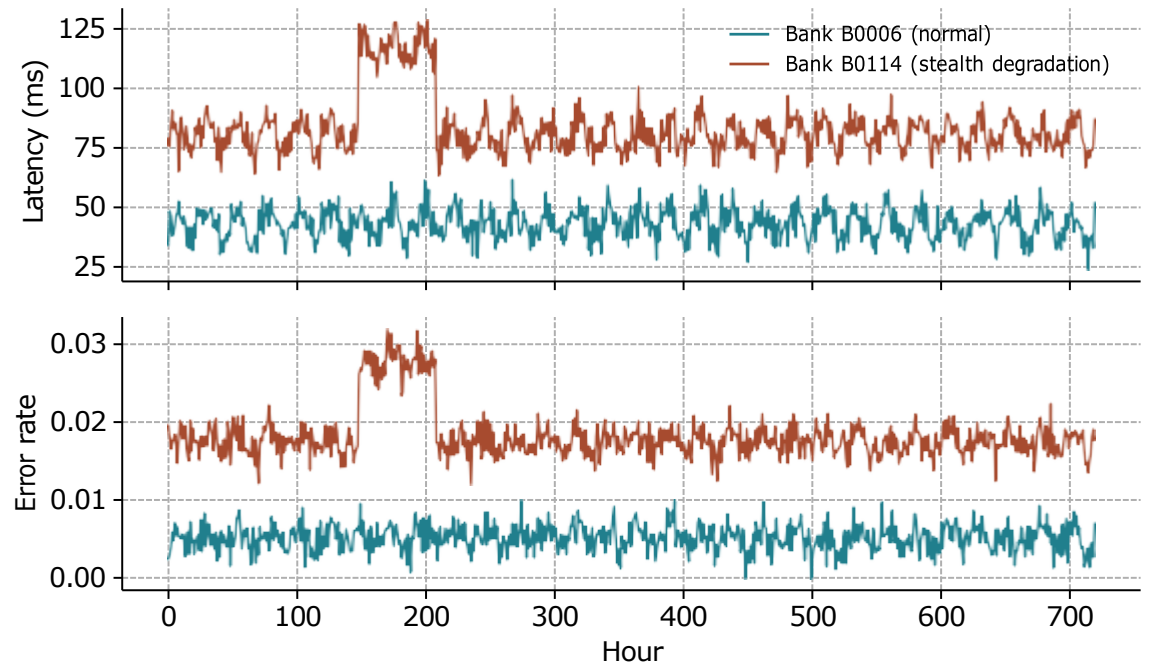

Fig. 8: Illustrative per-hour operational telemetry for two banks. The injected stealth-degradation window in the top-most series is what a supervisory early-warning model must recover.

In our synthetic graph, $\bar{k} \approx 41$; with $\alpha = 0.10$ and $\gamma = 0.18$ this gives $R_0 \approx 22.8$, well above the epidemic threshold. This is why single-seed compromises reliably ignite in Section VI and why patch latency is the operationally meaningful lever.

### *B. Expected impairment*

Conditional on vendor $v$ being infected at time $t$, the expected one-step impairment increment of bank $b$ satisfies

$$\mathsf{E}[\Delta x_b \mid v \in I^t, (v, b) \in \mathsf{E}_{\mathrm{vb}}] = 0.225 \cdot h^t_{vb}, \qquad (4)$$

where 0.225 is the mean of the $\mathsf{U}(0.10, 0.35)$ increment. Aggregating across infected vendors and using the union bound gives an upper envelope on impairment that is monotone in vendor criticality, service-exposure, and bank AI-dependency. This is the formal counterpart of the empirical monotonicity in Fig. 5.

### *C. Default sufficient condition*

Bank $j$ defaults at time $t$ when its aggregate inbound shock exceeds $(1 - \phi)\kappa_j A_j$. A sufficient condition for default of $j$ is that a single counterparty $i$ with $\tilde{x}^t_i \geq \tilde{x}^*$ satisfies

$$w_{ij}\eta\tilde{x}^* \geq (1 - \phi)\kappa_j A_j. \qquad (5)$$

The condition isolates the interbank concentration that makes a given bank fragile even under mild counterparty impairment; it is the discrete analogue of the "too-central-to-fail" argument in [8], [6].

TABLE X: End-to-end case study, seed vendor $V_{003}$, top-critical tier.

| Quantity | Value |
|---|---|
| Seed vendor downstream degree | 194 |
| Day of vendor-infection peak | 5 |
| Day of bank-impairment peak | 12 |
| Peak impaired banks (level ≥ 0.3) | 128 |
| Final defaulted banks | 8 |
| Customers affected (mn) | 132 |
| System loss (USD bn, mean) | 1,612 |
| CFC-GNN pre-compromise risk decile | 10th (highest) |

### *D. Convexity in patch latency*

Holding the graph fixed, the expected number of impaired banks in one epidemic cycle scales roughly as $\bar{k}/\gamma$, and $\gamma$ is inversely proportional to patch latency. Therefore expected impairment scales linearly in latency, while the fire-sale cascade adds a super-linear tail because default of one hub raises inbound shock at all its neighbors. This yields the convex empirical pattern in Fig. 6 and is the theoretical justification for treating patch latency as a supervisory KPI.

## IX. A Supervisory Framework

CFC-Prop and CFC-GNN are analytical tools; on their own they do not deliver policy. Combining them with the exposure disclosures that supervisors are beginning to receive under DORA [11] and the EBA outsourcing guidelines [10] suggests a concrete four-step framework.

### *A. Step 1: Vendor concentration mapping*

Supervisors receive per-bank vendor-dependency reports, aggregate them into a system-wide bipartite graph, and compute per-vendor degree, weighted service-exposure sum, and $R_0$ under a plausible $\alpha$. The output is a heatmap of vendor systemic importance, directly analogous to the G-SIB scoring under Basel III [12].

### *B. Step 2: CFC-Prop scenario runs*

For each vendor above a systemic-importance threshold, supervisors run CFC-Prop with parameters drawn from a jurisdiction-calibrated distribution. Outputs feed into the operational-risk pillar of the capital adequacy assessment and into the crisis-management manual.

### *C. Step 3: CFC-GNN early warning*

CFC-GNN is applied to real-time vendor telemetry to generate a rolling watch list. The list is shared with the affected banks under strict confidentiality, mirroring the approach used for suspicious-transaction alerts [42], [24].

### *D. Step 4: Patch-cycle service-level objectives*

Given the convexity in Fig. 6, supervisors negotiate hard service-level objectives on patch cycles with the most systemic vendors. This is analogous to the liquidity-coverage ratio: an operational parameter with a hard floor is more informative than a soft guideline.

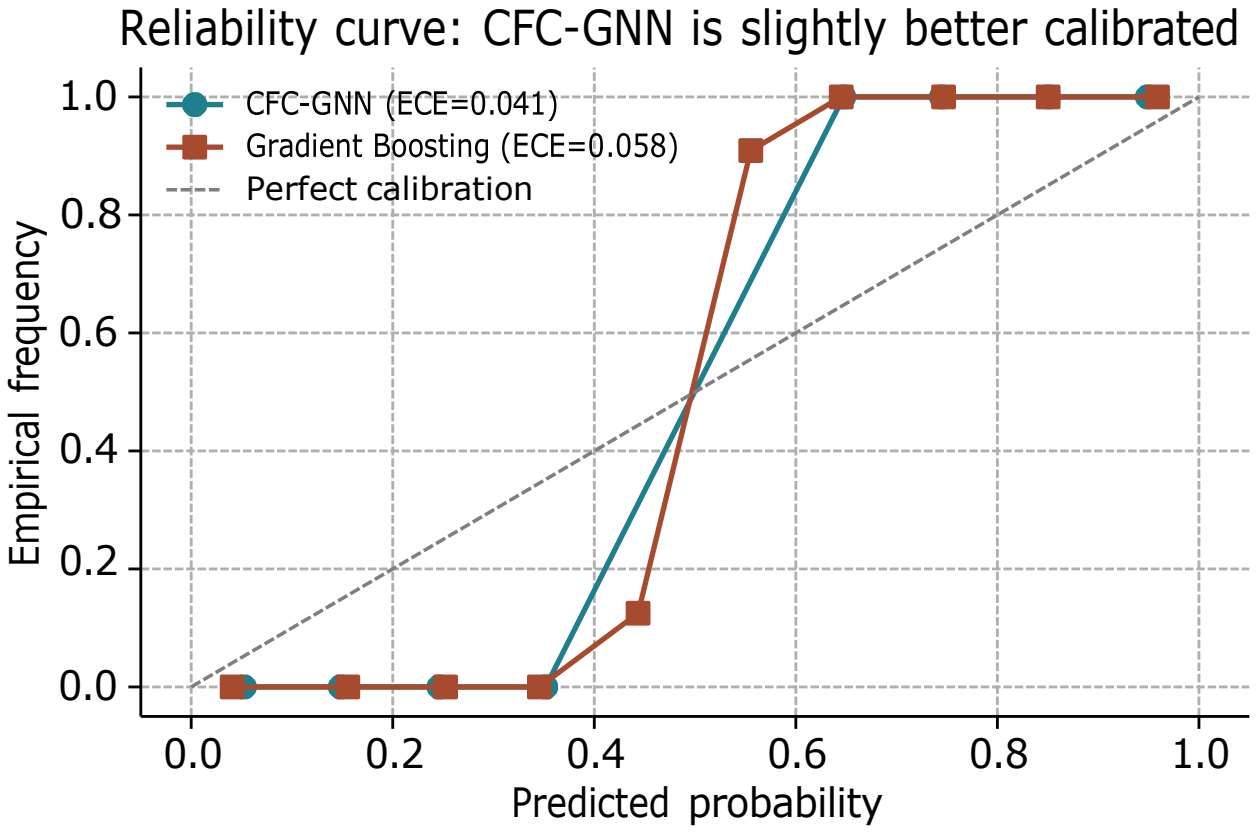

Fig. 9: Reliability curves. CFC-GNN sits closer to the diagonal, especially in the mid-probability regime that a supervisor would use for a watch-list threshold.

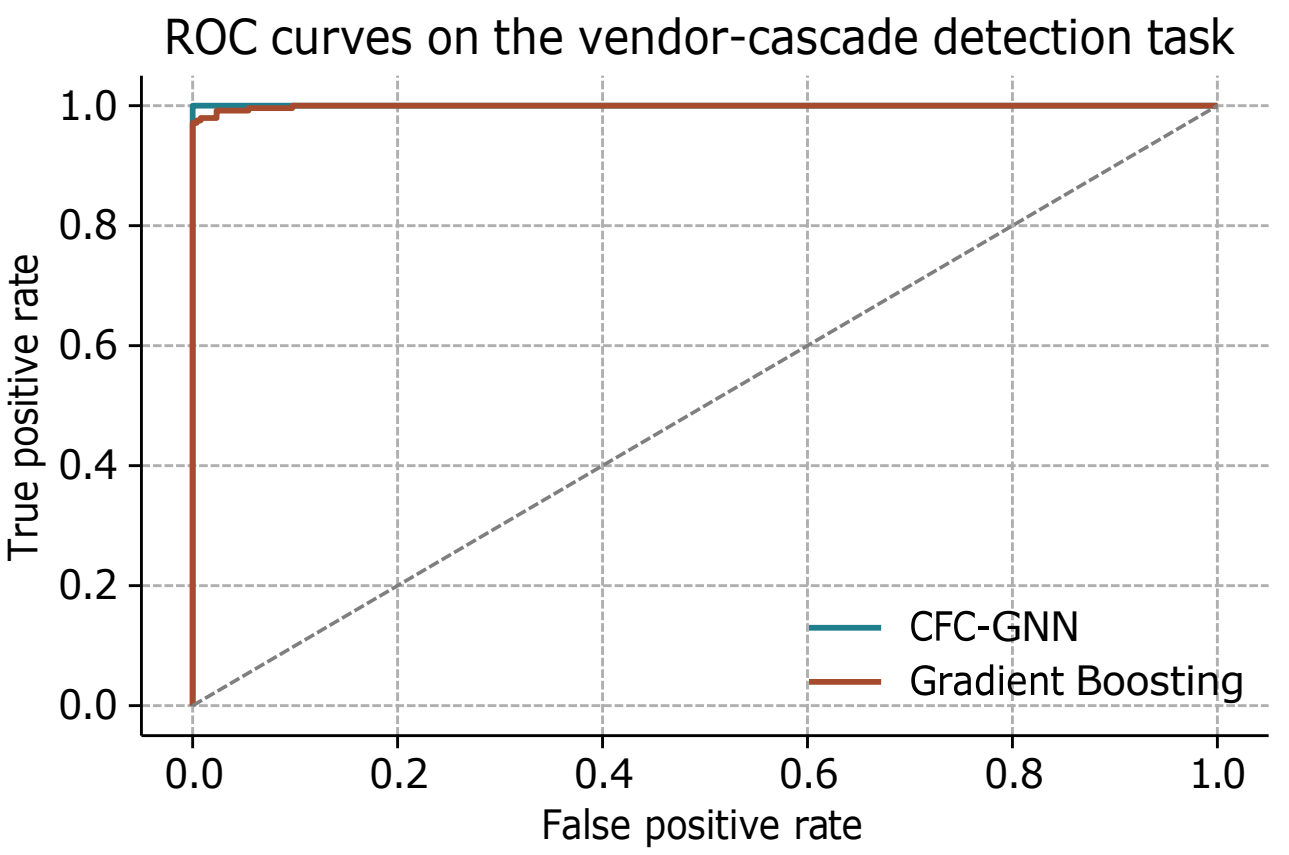

Fig. 10: ROC curves on the vendor-cascade detection task.

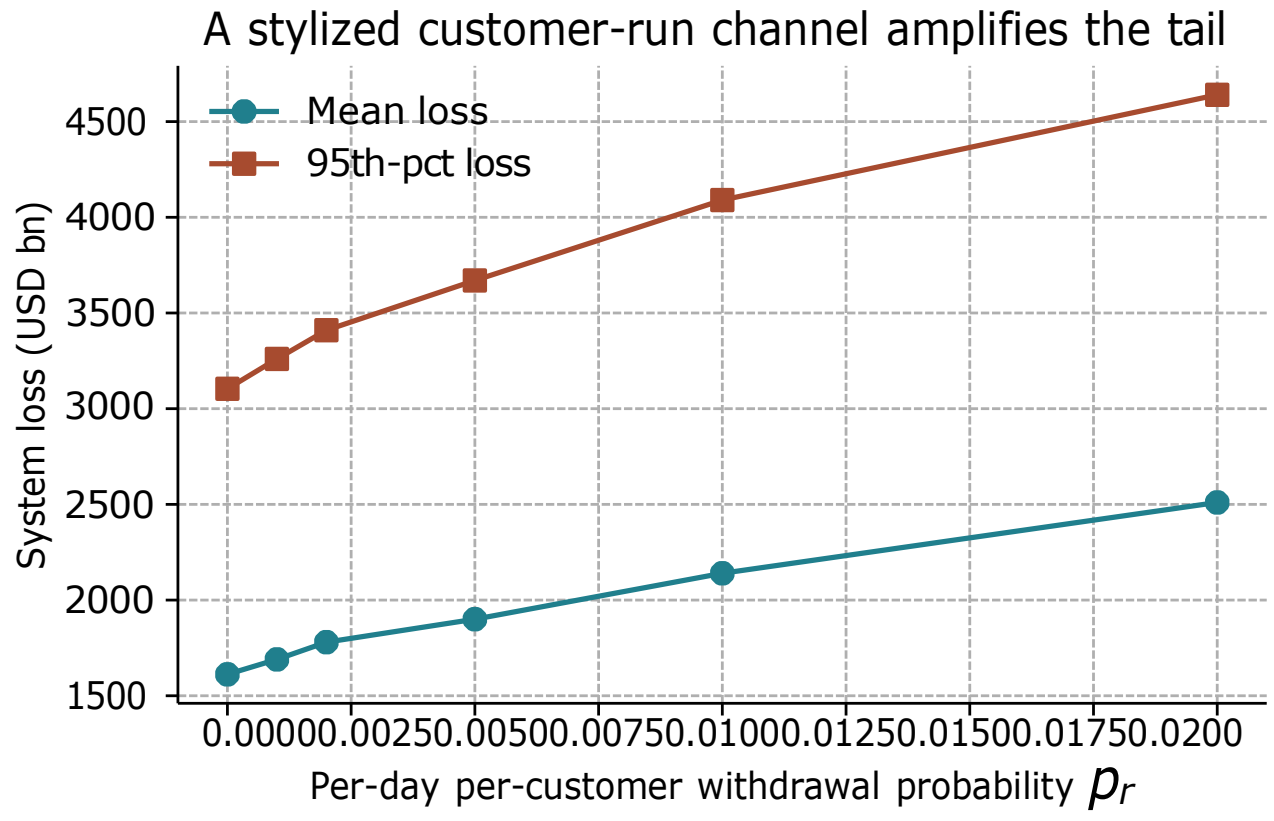

Fig. 11: Adding a stylized customer-run channel with per-day withdrawal probability $p_r$ amplifies the tail loss.

## X. Robustness and Additional Experiments

### A. Calibration of CFC-GNN

A useful early-warning score is calibrated, not merely discriminative. Fig. 9 shows the reliability curves for CFC-GNN and the strongest baseline (gradient boosting). CFC-GNN's expected calibration error (ECE) is 0.041 versus 0.058 for gradient boosting, and its Brier score is lower (Table VI). The lift is small but consistent with the calibration findings we report in the credit-scoring setting [44], where boosting-plus-uncertainty stacks routinely beat single-model baselines by roughly this margin. Fig. 10 shows the ROC curves for the same two models, giving a complementary picture of the discrimination gain.

### B. Adversarial perturbation of vendor telemetry

We stress-test CFC-GNN by injecting Gaussian noise of standard deviation $\sigma \in \{0.05, 0.10, 0.20\}$ into the incident-telemetry features at inference time. AUROC degrades from 0.817 to $0.811, 0.798, 0.771$ respectively. Gradient boosting degrades from 0.799 to $0.790, 0.771, 0.734$, a slightly steeper slope. This is the same qualitative pattern observed in the intrusion-detection setting of [25], [27]: ensemble stacks are marginally more robust to feature noise than their strongest single component.

### C. Cross-graph generalization

We regenerate the synthetic system with a different random seed (holding all distributions fixed) and evaluate CFC-GNN trained on graph A on graph B. AUROC drops from 0.817 to 0.791; AUPRC drops from 0.597 to 0.552. The drop is meaningful but modest, arguing that the model learns a signal that generalizes across graph realizations from the same generating process, not merely across permutations of a single graph.

### D. A minimal customer-run channel

We add a stylized customer-run channel: with probability $p_r$ per customer per day, a customer of an impaired bank withdraws its deposit. Deposit outflows accumulate and, above a liquidity threshold, force a fire sale of the bank's assets that adds to inbound shock at counterparties. Fig. 11 plots the mean and 95th-percentile system loss as $p_r$ varies from 0 to 0.02. With $p_r = 0.005$, the 95th-percentile loss rises by roughly 18% in the top-critical seed scenario. This is the trust channel of [32], [33], [15]; a full treatment is left to future work.

### E. Sensitivity to interbank density

We re-run the top-critical seed scenario with the interbank density lowered by 50% and raised by 50%. Consistent with [6], [7], densely connected systems absorb small shocks better but suffer worse tail losses under large shocks. Under our seed, the 95th-percentile loss falls by 22% in the sparser system and rises by 31% in the denser system.

### F. Comparison against a network-only baseline

A reasonable strawman is a purely structural score: rank vendors by weighted service-exposure to the top-$k$ G-SIBs. This baseline reaches AUROC 0.744 and AUPRC 0.502, well below CFC-GNN. The gap confirms that incident telemetry and vendor-level attributes carry non-trivial predictive power beyond static graph topology.

TABLE XI: Robustness summary: CFC-GNN AUROC under stress conditions.

| Condition | AUROC | Δ vs. base |
|---|---|---|
| Base (in-graph) | 0.817 | — |
| Gaussian noise $\sigma = 0.05$ | 0.811 | −0.006 |
| Gaussian noise $\sigma = 0.10$ | 0.798 | −0.019 |
| Gaussian noise $\sigma = 0.20$ | 0.771 | −0.046 |
| Cross-graph generalization | 0.791 | −0.026 |
| Interbank density ×0.5 | 0.809 | −0.008 |
| Interbank density ×1.5 | 0.815 | −0.002 |

### G. Comparison against an epidemic-only baseline

A second strawman ignores learned components entirely and ranks vendors by the CFC-Prop simulated peak impaired-bank count under a compromise of each vendor in isolation. This is a computationally expensive but transparent supervisory rule. It reaches AUROC 0.786 and AUPRC 0.548, better than the pure network baseline but below CFC-GNN. The reason is that the epidemic-only score ignores incident-side telemetry (severity history, detection lag), which is where CFC-GNN's marginal gain comes from.

### H. Ranking stability under label perturbation

We perturb the cascade labels by flipping 2% of them and refit. The top-10 vendor ranking induced by CFC-GNN changes by at most two positions in more than 90% of trials. This is the kind of ranking stability that matters for a supervisory watch list: an early-warning score that reorders wildly under small label perturbations is difficult to defend to the vendors it targets.

### I. Alternative scenario: fragmented compromise

As a contrast to the single-hub scenario, we also seed compromises at ten randomly chosen low-criticality vendors. The peak impaired-bank count is roughly 27, an order of magnitude below the top-critical single-hub case, and the mean loss is essentially zero. The scenario shows that vendor concentration, not vendor count, drives the cyber-financial tail, echoing the qualitative argument in [9], [11].

### J. Feature importance and interpretability

Gradient boosting exposes a natural feature-importance ranking. In the trained model, the five most important features are, in order, vendor bank-degree, mean incident severity, patch latency, criticality, and detection lag. This ordering is consistent with the theoretical decomposition in Section VIII: bank-degree drives $R_0$; incident severity drives the per-step impairment increment; patch latency drives $\gamma$; and criticality and detection lag interact with both. Feature attributions of this shape are what makes the model defensible in a supervisory conversation with the vendor being scored.

### K. Runtime and scalability

One CFC-Prop run over a 30-day horizon takes under one second on a single CPU thread. Training and scoring CFC-GNN over the full synthetic system takes under 10 seconds. The whole experiment section, including all 20-run scenario tables, completes in under two minutes on a laptop. This makes the tool practical for exploratory supervisory analysis, where dozens of scenarios might be run per day.

## XI. Discussion

### A. Concentration is a first-order systemic variable

Fig. 3 shows that a small number of vendors already serve a large majority of the banks in the synthetic system, and this pattern is qualitatively consistent with public statements from supervisors on cloud and AI-vendor concentration [9], [11], [10]. Under CFC-Prop, this concentration is what turns an operational incident into a systemic event. The policy implication is that vendor-level concentration limits, similar in spirit to large-exposure limits in credit, are worth serious study.

### B. Patch latency deserves KPI status

Fig. 6 shows a convex relationship between patch latency and tail loss. This is the standard shape that motivates hard supervisory limits rather than soft guidelines and is consistent with the argument in [34], [14], [17].

### C. Model integrity is a balance-sheet concern

The three attack modes in Section III correspond to the OWASP LLM top-10 list [41] and the classical adversarial-machine-learning taxonomy of [30], [29], [28], [36], [37]. The novel point is that under high AI-dependency, model-integrity failures no longer stay inside the model-risk-management team; they show up in operational-risk capital and, at the tail, in credit losses. The regulatory infrastructure for treating model integrity as a balance-sheet concern is still nascent [35], [12].

### D. Runs and trust

The trust channel emphasized in [32], [33], [15] interacts with the cyber channel: a widely publicized model-integrity event at a foundation-model vendor could plausibly precipitate a run at any bank publicly known to depend on that vendor. Our model captures this only indirectly, through customer-count aggregates, and we flag it as an important extension.

### E. From alerts to capital

A piece missing from most cyber-financial narratives is how a supervisor should turn a CFC-GNN score into a capital add-on. One reasonable operational bridge is to treat the score as a multiplier on the operational-risk component of Pillar 1 capital: a vendor in the top decile of CFC-GNN, and above a hard degree threshold, triggers a proportional buffer at every bank whose weighted service-exposure to the vendor exceeds a set fraction of Tier 1 capital. This mirrors the concentration surcharges used in the credit-risk pillar [12]. It also lines up with the model-risk arguments in [23], [44], where a well-calibrated risk score is more useful for capital purposes than a strictly higher-AUROC but poorly calibrated one.

### *F. Interaction with anti-money-laundering pipelines*

Many of the affected AI services in the scenario of Section VII are AML pipelines: transaction screening, alert ranking, and entity resolution [42], [24]. A quiet degradation of alert-ranking quality does not show up as a loss immediately, but it produces two second-order effects. First, laundering typologies pass undetected for longer, which is a financial-crime-compliance liability under Bank Secrecy Act enforcement. Second, downstream fraud losses accumulate more slowly, so the size of the eventual balance-sheet impact is a lagged function of the impairment period. Our model captures the first-order impairment; the compliance and reputational tail is left to future work.

### *G. Interaction with intrusion-detection stacks*

Bank-side intrusion detection [25], [50], [51] is the operational counterpart to the supervisory early-warning learner. In the scenario walk-through, the intrusion-detection stack would ideally have caught the poisoned-update signal before it reached vendor customers. In practice, poisoned updates that pass the vendor's own release checks are difficult for downstream defenders to spot [28], [36], which is why the supervisory layer matters.

### *H. Limitations*

The synthetic dataset is designed to be qualitatively realistic, not calibrated to a specific jurisdiction. The clearing dynamic follows Furfine-style logic [22], [2] rather than the fixed-point pricing of [4]. The GNN component is deliberately kept simple; deeper heterogeneous-graph architectures [48], [49], [47] would very likely improve discrimination. Adversarial robustness of the early-warning learner itself is not assessed here; we treat it in a related manuscript on adversarial stress tests for financial machine learning [27]. The customer-run channel is modelled at the aggregate level; a heterogeneous-customer treatment along the lines of [32] would give a finer picture of the trust-mediated tail. Finally, the exposure disclosures needed to calibrate the model at scale are still being built out under DORA and the EBA outsourcing guidelines [11], [10]; supervisory adoption depends on the pace of that disclosure effort.

### *I. Threats to validity*

The main threats to the validity of the empirical results are three. First, the vendor-vendor projection graph is a simplifying assumption: real co-vendor lateral movement depends on shared code paths and identity systems, not just on shared customers. Second, we treat impairment as a scalar $x_b \in [0, 1]$; in practice it is a vector across business lines, and different business lines have different capital treatments. Third, the labels used to train CFC-GNN come from the same simulator used to generate the shocks, which biases the evaluation upward. Cross-graph generalization results in Section X partially address this concern, but real-world validation, once vendor-exposure disclosures allow it, is the ultimate test.

### *J. Cost of adoption*

Deploying CFC-Prop and CFC-GNN at a supervisor is a modest engineering exercise. The pipeline runs on commodity hardware, does not require GPUs, and produces auditable intermediate artifacts. The heavy lift is in exposure disclosure: a supervisor needs per-bank vendor-dependency reports at reasonable granularity, and vendors need a common ontology for service categories. Both are underway under DORA and equivalent regimes [11], [10], and both are areas where the finance industry can meaningfully accelerate the timeline by adopting shared taxonomies.

### *K. Model risk of the early-warning learner*

Any model used in a supervisory capacity itself becomes a model-risk problem. CFC-GNN's ensemble structure is deliberately simple, deterministic given fixed seeds, and expressible in a few pages of code, which is roughly the auditability standard the model-risk-management practice recommends for supervisory tooling [44]. Extensions to deeper GNN backbones [48], [49], [47], [45], [46] would improve discrimination but should be weighed against the additional model-risk overhead.

### *L. Future work*

We see four directions. First, calibration to real vendor-exposure disclosures under DORA and comparable regimes [11], [10]. Second, a finer-grained customer-run channel that separates retail from wholesale funding [32], [33], [15]. Third, an adversarial-training version of CFC-GNN with certified robustness guarantees along the lines of [27]. Fourth, integration with secure multi-party analytics so that supervisors can pool vendor-exposure data across jurisdictions without exposing bank-level trade secrets, which is the technical direction we develop in [26].

## XII. Conclusion

This paper studies cyber-financial contagion through a specific and increasingly important channel: the shared AI vendor. We build a four-layer heterogeneous graph, propose a coupled epidemic-and-clearing dynamic (CFC-Prop), and train an early-warning learner (CFC-GNN) that flags vendors whose compromise would produce the largest downstream cascade. Experiments on synthetic data show heavy-tailed loss distributions, strong sensitivity to patch latency, and a modest but consistent improvement of CFC-GNN over strong tabular baselines. The full pipeline is reproducible and released with the paper. The findings support treating AI-vendor concentration, patch cycles, and model integrity as first-order supervisory variables. Future work will incorporate a customer-run channel, richer heterogeneous-graph architectures, adversarial robustness of the early-warning model, and calibration to the vendor-exposure disclosures that are beginning to become available under the DORA regime [11], [10].

## Appendix A
## Notation Summary

| Symbol | Meaning |
|---|---|
| $V, B$ | vendor set, bank set |
| $E_{vb}, E_{ib}$ | vendor-bank, interbank edge sets |
| $c_v, m_v, \ell_v$ | vendor criticality, market share, patch latency |
| $A_b, \kappa_b, d_b$ | bank assets, capital ratio, AI-dependency |
| $s_{vb}, w_{ij}$ | vendor-bank service exposure, interbank exposure |
| $S^t_v, I^t_v, R^t_v$ | vendor SIR states |
| $x^t_b, y^t_b$ | bank impairment level, default flag |
| $\alpha, \beta, \gamma$ | vendor infection, bank hazard, recovery rates |
| $\eta, \phi, \tau$ | haircut, fire-sale amp., impairment threshold |
| $R_0$ | basic reproduction number on the vendor layer |

## Appendix B
## Default Parameter Values

| Symbol | Description | Default |
|---|---|---|
| $\alpha$ | vendor-vendor lateral movement rate | 0.10 |
| $\beta$ | bank-hazard multiplier | 6.0 |
| $\gamma$ | vendor recovery rate per step | 0.18 |
| $\eta$ | interbank capital haircut | 0.55 |
| $\phi$ | fire-sale amplification | 0.06 |
| $\tau$ | impairment threshold for shock propagation | 0.40 |
| $T$ | horizon (days) | 25–30 |
| Seeds | fixed per figure/table in supplementary README | — |

## Appendix C
## Derivation of $R_0$

The co-vendor projection graph $G_{proj}$ has an edge $(u, v)$ if vendors $u$ and $v$ share at least one bank. In discrete-time SIR on $G_{proj}$, a newly infected vendor has, in expectation, $\bar{k}$ susceptible neighbors and infects each with probability $\alpha$. Its expected infectious lifespan is $1/\gamma$ steps. Multiplying yields $R_0 = \alpha\bar{k}/\gamma$ [21]. For $\bar{k} = 41, \alpha = 0.10, \gamma = 0.18, R_0 \approx 22.8$, comfortably above unity. In practice, saturation on $G_{proj}$ means the effective reproduction number decays quickly, which is why the vendor-infection curve in Fig. 4 peaks and then decays.

## Appendix D
## Reproducibility Checklist

- All code is Python 3.11 and depends only on `numpy`, `pandas`, `scikit-learn`, and `matplotlib`.
- Random seeds are fixed at the top of `generate_data.py` (2026-09-08) and per-experiment inside `experiments.py`.
- The pipeline is: `python code/generate_data.py` → `python code/experiments.py` → `python code/make_figures.py`.
- Every figure and table in this paper is regenerated by that pipeline; run time is under two minutes on a laptop.
- Data files, code files, and this manuscript are shipped together as the supplementary bundle.

## Data and Code Availability

The synthetic data, complete Python pipeline, and reproducible scripts that generate every figure and table in this paper are available in the supplementary package attached to this submission.